\documentclass[conference]{IEEEtran}
\IEEEoverridecommandlockouts

\usepackage{cite}
\usepackage{subcaption}
\usepackage{amsmath,amssymb,amsfonts}
\usepackage{algorithmic}
\usepackage{graphicx}
\usepackage{textcomp}
\usepackage{xcolor}
\usepackage{orcidlink}
\usepackage{hyperref}
\def\BibTeX{{\rm B\kern-.05em{\sc i\kern-.025em b}\kern-.08em
    T\kern-.1667em\lower.7ex\hbox{E}\kern-.125emX}}

\title{TGO-II: Representational Similarity Observatory}
\author{
\IEEEauthorblockN{Kaustubh Kapil\orcidlink{0009-0000-4918-8452} and Kishor P. Upla\orcidlink{0000-0001-6306-0682}}
\IEEEauthorblockA{
Department of Electronics Engineering\\
Sardar Vallabhai National Institute of Technology (SVNIT), Surat, India\\
u24ec049@eced.svnit.ac.in, kishorupla@gmail.com
}
}
\begin{document}

\maketitle
\begin{abstract}

While Vision Transformers have achieved remarkable success across computer vision and language applications, the geometric evolution of their internal representations throughout training remains insufficiently understood. Existing analyses primarily focus on attention mechanisms and downstream performance, leaving the evolution of representation geometry largely unexplored. In this work, we present \textbf{Transformer Geometry Observatory-II (TGO-II)}, a representation geometry analysis framework designed to investigate how transformer representations evolve during supervised training. TGO-II analyzes Vision Transformer (ViT-Small/16) representations using Centered Kernel Alignment (CKA), Singular Vector Canonical Correlation Analysis (SVCCA), Two-Nearest Neighbor Intrinsic Dimensionality (TwoNN-ID), and token covariance analysis. Our experiments reveal three key observations. First, both CKA and SVCCA exhibit a progressive decline throughout training, indicating increasing representational specialization across transformer layers. Second, intrinsic dimensionality consistently increases before stabilizing, suggesting progressive expansion of the representation manifold into a larger set of locally accessible degrees of freedom. Third, token covariance and coupling analyses demonstrate that strong token interaction structure persists throughout training, challenging the hypothesis that increasing representational complexity arises primarily from progressive token independence. These findings suggest that representation complexity and layer specialization emerge simultaneously during training. Furthermore, the observed growth in intrinsic dimensionality appears to be accompanied by manifold expansion rather than token decoupling. Together, these observations motivate a new hypothesis in which Vision Transformers increase representational complexity through progressively richer transformations while preserving strong token interaction structure.

\end{abstract}

\section{Introduction}

Vision Transformers (ViTs) have rapidly become one of the dominant architectures in modern machine learning, achieving state-of-the-art performance across computer vision, language modeling, multimodal reasoning, and generative modeling tasks. While their empirical capabilities have been extensively studied, the geometric evolution of their internal representations during training remains comparatively less understood. Most existing analyses focus on architectural components such as self-attention mechanisms, feed-forward networks, token interactions, or downstream performance metrics. These studies have provided valuable insights into transformer operation; however, they often investigate individual mechanisms in isolation rather than the geometry of the learned representations themselves. As a result, fundamental questions regarding how representations evolve, specialize, and organize throughout training remain largely unanswered.

Transformer Geometry Observatory (TGO) was proposed as a systematic framework for studying transformer learning dynamics through a collection of geometry-focused observatories. The first observatory, \textbf{TGO-I: Spectral Geometry Observatory}~\cite{tgoi}, investigated the covariance structure of Vision Transformer representations through eigenspectra, Effective Rank, Spectral Entropy, and Spectral Anisotropy analyses. It revealed a consistent increase in dimensional utilization throughout training, accompanied by decreasing anisotropy and increasing spectral entropy. Furthermore, several groups of adjacent layers exhibited remarkably similar spectral trajectories, motivating the hypothesis that portions of the network may exhibit representational redundancy. However, spectral observables alone are insufficient for determining whether spectrally similar layers perform similar computations. Two layers may exhibit nearly identical covariance spectra while occupying entirely different representational subspaces. Consequently, the observations of TGO-I motivate a deeper investigation into representation geometry itself.

To address this question, we introduce \textbf{TGO-II: Representation Geometry Observatory}. Rather than studying how variance is distributed across feature directions, TGO-II investigates how representations relate to one another throughout training. Specifically, we analyze representation similarity using Centered Kernel Alignment (CKA)~\cite{cka} and Singular Vector Canonical Correlation Analysis (SVCCA)~\cite{svcca}, estimate representation manifold complexity through Two-Nearest Neighbor Intrinsic Dimensionality (TwoNN-ID)~\cite{twonn}, and examine token interaction structure through token covariance and token coupling analyses. Our analyses reveal that representational similarity progressively decreases throughout training despite the spectral clustering observed in TGO-I. Simultaneously, intrinsic dimensionality consistently increases before stabilizing, suggesting progressive expansion of the representation manifold. Furthermore, token covariance analyses indicate that strong token interaction structure persists throughout training, challenging the hypothesis that increasing representational complexity arises primarily through token decoupling. Taken together, these observations suggest that transformer training simultaneously increases representational complexity while driving layer specialization. These findings motivate a new working hypothesis in which complexity growth emerges through progressively richer transformations of highly coupled token systems rather than through progressive token independence.

The contributions of this work are summarized as follows:

\begin{itemize}
\item This study introduces TGO-II, a representation geometry observatory for studying transformer learning dynamics.
\item Representation similarity has been analyzed throughout training using CKA and SVCCA.
\item Investigation into representation manifold evolution through TwoNN intrinsic dimensionality estimation.
\item We study token interaction structure through token covariance and token coupling analyses.
\item Empirical evidence that increasing representational complexity can coexist with decreasing inter-layer similarity and persistent token coupling.
\end{itemize}

\begin{figure*}[t]
    \centering

    \subfloat[Mean CKA over training\label{fig:mean_cka}]{
        \includegraphics[width=0.48\linewidth]{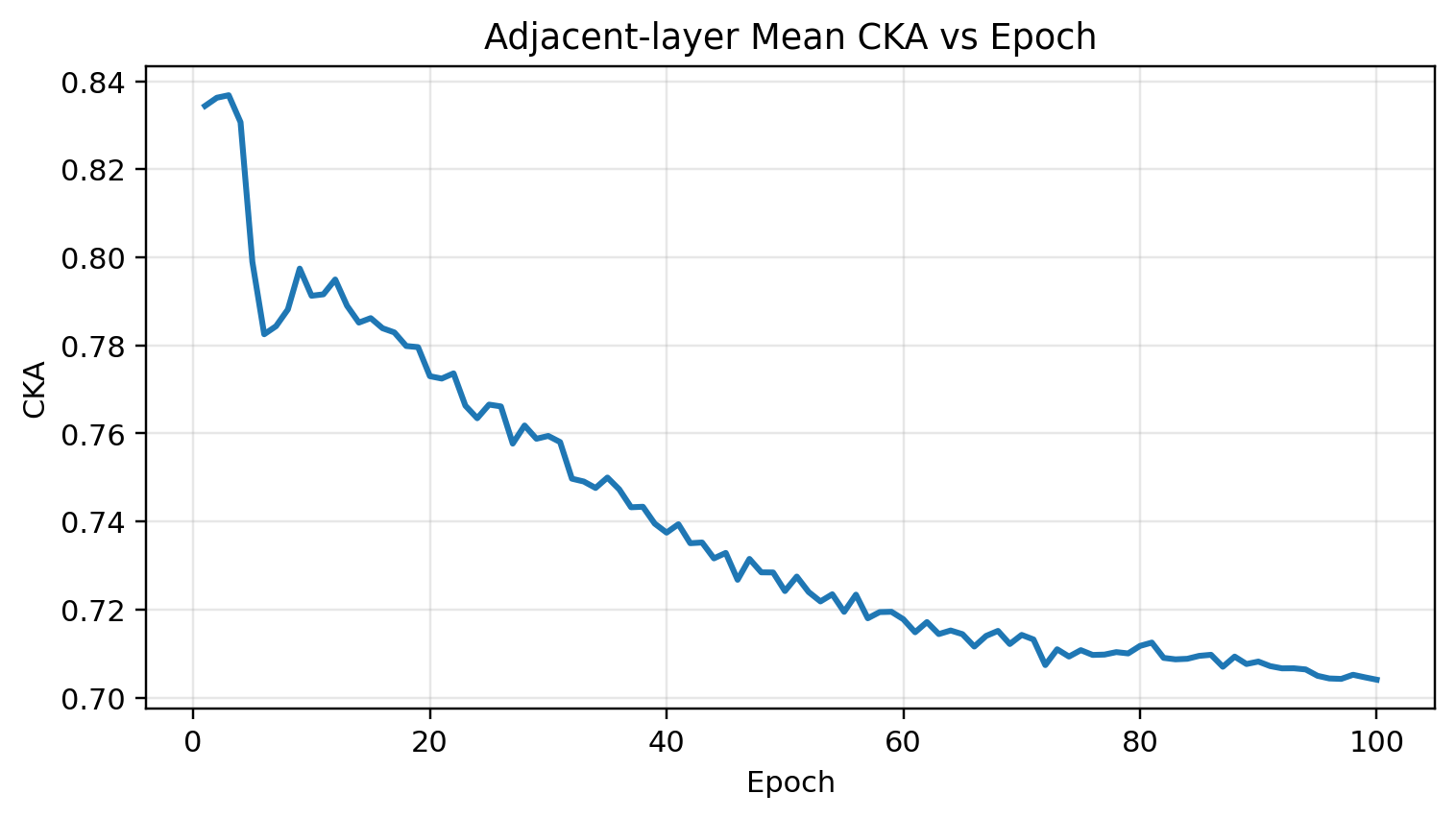}
    }
    \hfill
    \subfloat[Adjacent-layer CKA over training\label{fig:adj_cka}]{
        \includegraphics[width=0.48\linewidth]{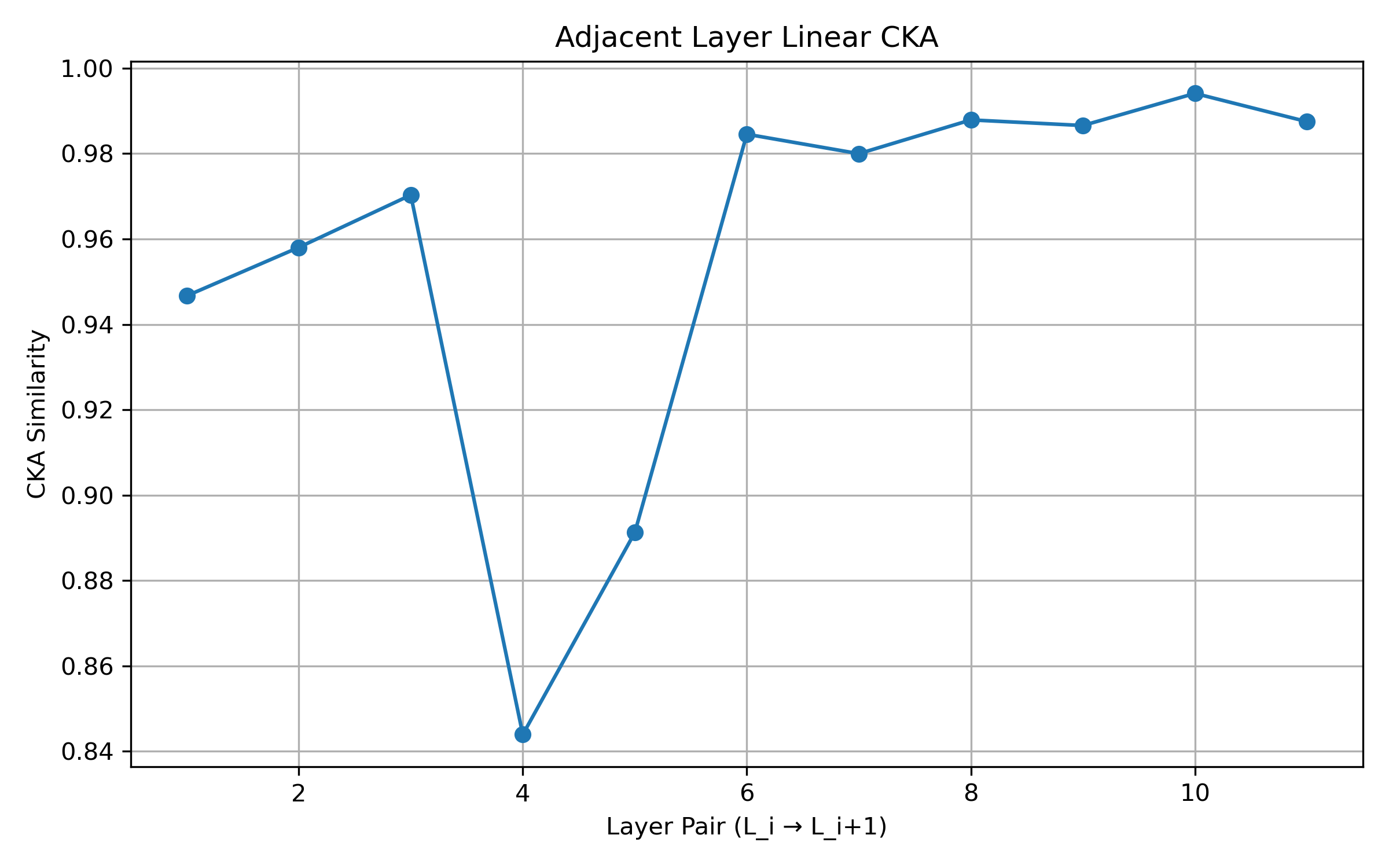}
    }

    \caption{Evolution of CKA throughout training. The left figure shows the mean CKA across all layer pairs, while the right figure shows the mean adjacent-layer CKA.}
    \label{fig:cka_combined}
\end{figure*}

\subsection{Research Questions}

The primary objective of TGO-II is to investigate three questions:

\begin{enumerate}
\item Do spectrally similar layers correspond to representationally similar layers?
\item How does the intrinsic dimensionality of transformer representations evolve throughout training?
\item Is increasing representational complexity driven by progressive token independence or by increasingly sophisticated transformations of strongly coupled token systems?
\end{enumerate}
The complete implementation of TGO-II, trained checkpoints, observatory pipelines,
and generated artifacts are publicly available at:
\href{https://github.com/KaustubhKapil/Transformer_Geometry_Observatory_Part-2}{GitHub Repository}.

\begin{figure*}[t]
    \centering

    \subfloat[Mean SVCCA over training\label{fig:mean_svcca}]{
        \includegraphics[width=0.48\linewidth]{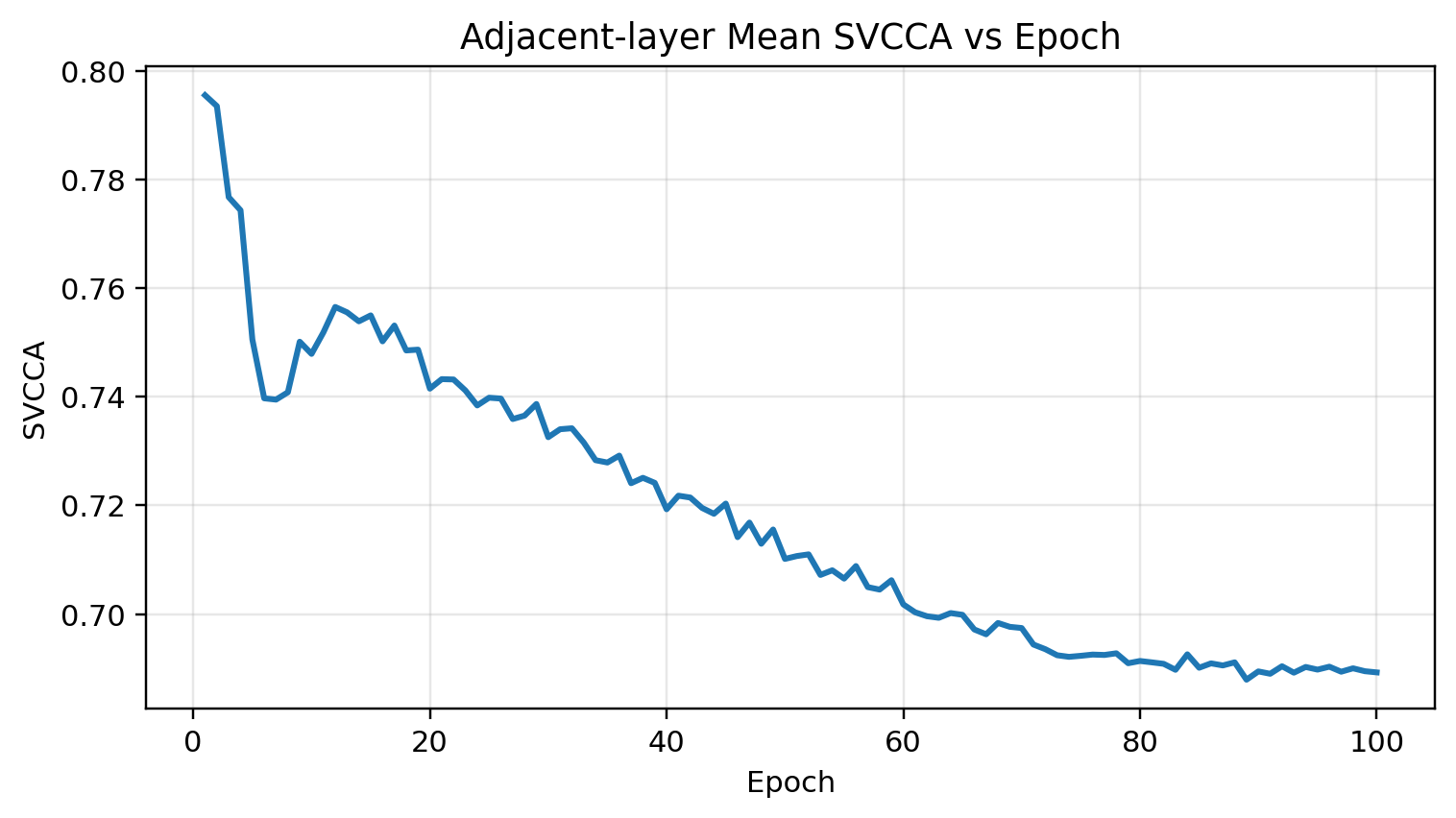}
    }
    \hfill
    \subfloat[Adjacent-layer SVCCA over Transformer Depth\label{fig:adj_svcca}]{
        \includegraphics[width=0.48\linewidth]{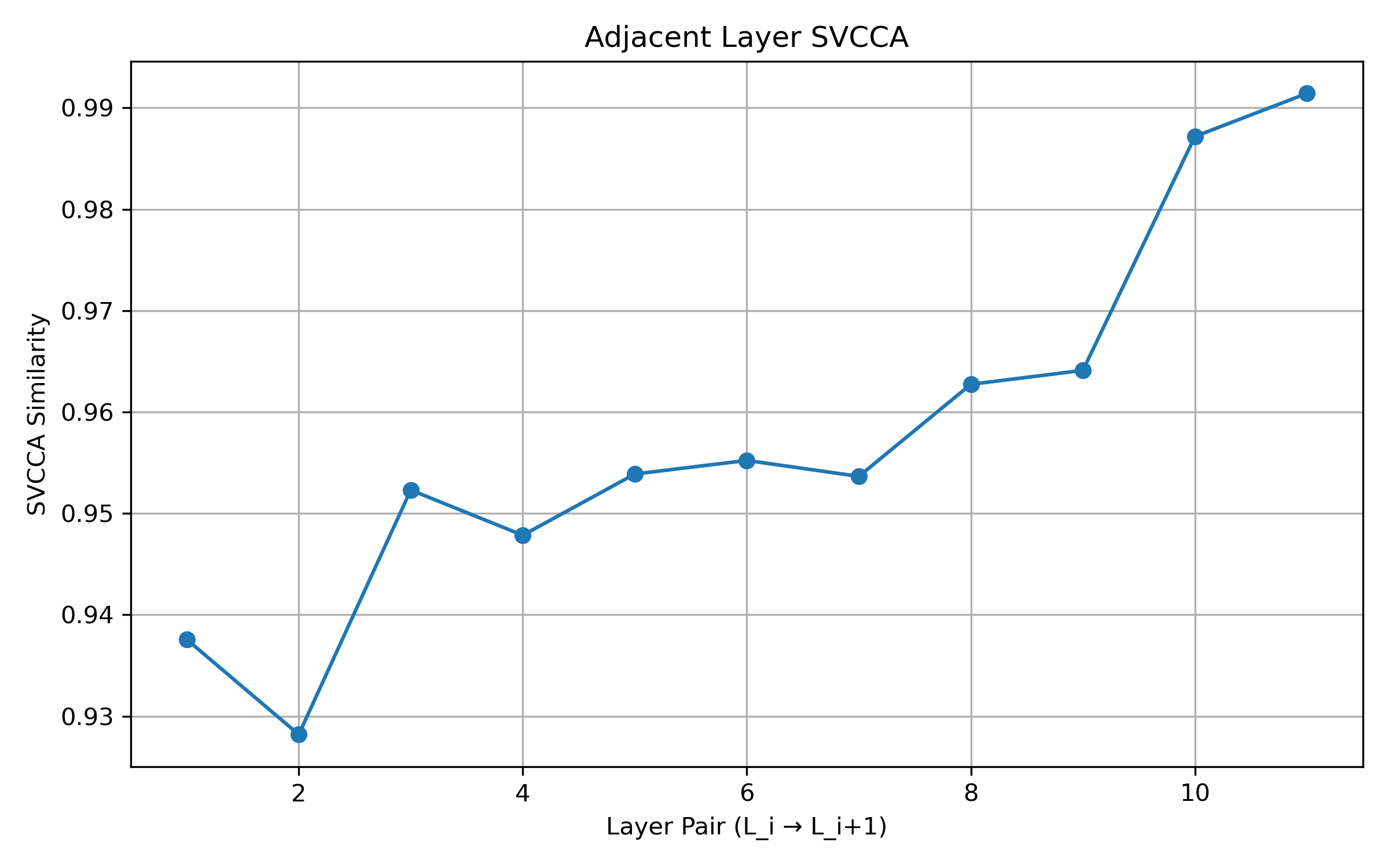}
    }

    \caption{a) Shows how the mean SVCCA shows a sudden drop and then stabilizes over the training schedule.
    b) Adjacent Layer SVCCA shows how the representational subspace similarity varies with the depth of the transformer}
    \label{fig:svcca_combined}
\end{figure*}

\section{Methodology and Representation Metrics}

The objective of TGO-II is to characterize the evolution of representation geometry throughout Vision Transformer training. To this end, we analyze representation similarity, intrinsic dimensionality, and token interaction structure across all Transformer layers of a ViT-Small/16~\cite{vit} model trained on ImageNet-100~\cite{imagenet} for 100 epochs. Activations were extracted from all Transformer blocks and the final CLS representation using forward hooks. To ensure consistent observability throughout training, all measurements were performed on a fixed analysis subset consisting of 1000 validation images.

For a given layer $l$, the representation matrix is defined as

\begin{equation}
\mathbf{H}_{l}
=
\begin{bmatrix}
\mathbf{h}_{l}(x_{1}) \\
\mathbf{h}_{l}(x_{2}) \\
\vdots \\
\mathbf{h}_{l}(x_{N})
\end{bmatrix}
\in
\mathbb{R}^{N \times D},
\end{equation}
where $N$ denotes the number of samples and $D$ denotes the embedding dimension. Each row corresponds to the CLS representation produced by layer $l$ for a particular input sample.

\subsection{Representation Similarity}
Representation similarity between layers is quantified using Centered Kernel Alignment (CKA)~\cite{cka} and Singular Vector Canonical Correlation Analysis (SVCCA)~\cite{svcca}. Given two representation matrices $\mathbf{H}_{i}$ and $\mathbf{H}_{j}$, linear CKA is computed as

\begin{equation}
\mathrm{CKA}
(\mathbf{H}_{i},\mathbf{H}_{j})
=
\frac{
\left\|
\mathbf{H}_{i}^{T}\mathbf{H}_{j}
\right\|_{F}^{2}
}
{
\left\|
\mathbf{H}_{i}^{T}\mathbf{H}_{i}
\right\|_{F}
\left\|
\mathbf{H}_{j}^{T}\mathbf{H}_{j}
\right\|_{F}
}.
\end{equation}

CKA measures the similarity of the geometric arrangement of samples within representation space and provides a scale-invariant estimate of representational alignment. SVCCA is computed by first projecting both representations onto their dominant singular subspaces and subsequently applying Canonical Correlation Analysis. The resulting score measures the degree of alignment between the principal representational subspaces of two layers. Together, CKA and SVCCA provide complementary measurements of representational similarity throughout training.

\subsection{Intrinsic Dimensionality}
To estimate the dimensionality of the learned representation manifold, TGO-II employs the Two-Nearest Neighbor Intrinsic Dimension estimator (TwoNN)~\cite{twonn}. For each representation vector $\mathbf{h}_{i}$, the distances to its first and second nearest neighbors are computed,

\begin{equation}
r_{1}(i)
\quad \text{and} \quad
r_{2}(i),
\end{equation}

and the ratio

\begin{equation}
\mu_i
=
\frac{r_{2}(i)}
     {r_{1}(i)}.
\end{equation}

The intrinsic dimensionality is subsequently estimated from the distribution of these ratios. TwoNN provides a local estimate of manifold complexity and quantifies the number of accessible degrees of freedom within representation space.

\subsection{Token Interaction Structure}
To investigate token relationships, token representations within each layer are analyzed through token covariance matrices. Let $\mathbf{T}_{l} \in \mathbb{R}^{T \times D}$ denote the token representation matrix at layer $l$, where $T$ is the number of tokens and $D$ is the embedding dimension. The token covariance matrix is computed as
\begin{equation}
\mathbf{C}^{\text{token}}_{l}
=
\frac{1}{D-1}
\left(
\mathbf{T}_{l}
-
\bar{\mathbf{T}}_{l}
\right)
\left(
\mathbf{T}_{l}
-
\bar{\mathbf{T}}_{l}
\right)^{T},
\end{equation}
where
\begin{equation}
\bar{\mathbf{T}}_{l}
=
\frac{1}{T}
\sum_{i=1}^{T}
\mathbf{T}_{l}^{(i)}
\end{equation}
denotes the feature-wise token mean. This matrix characterizes pairwise token relationships and reveals the degree to which token representations remain coupled throughout training. To quantify this interaction structure, a token coupling ratio is defined as
\begin{equation}
\mathrm{Coupling}
=
\frac{
\sum_{i \neq j}
|C_{ij}|
}
{
\sum_{i,j}
|C_{ij}|
}.
\end{equation}

These values approaching zero indicate predominantly independent token representations, while larger values indicate stronger token interaction structure. The observables studied in TGO-II therefore consist of CKA, SVCCA, TwoNN Intrinsic Dimensionality, Token Covariance, and Token Coupling. Collectively, these measurements characterize representational similarity, manifold complexity, and token interaction dynamics throughout training.

\section{Experiments}
This section describes the experimental configuration used to investigate the spectral evolution of Vision Transformer representations throughout training. A ViT-Small/16 model~\cite{vit} was trained on the ImageNet-100~\cite{imagenet}dataset for 100 epochs using a NVIDIA Quadro RTX 6000 GPU. To ensure consistent observability, all representational measurements were performed on a fixed analysis subset of 1000 validation images throughout training. The following subsections describe the model configuration, activation extraction pipeline, analysis subsets, and training parameters used throughout TGO-II.

This observatory emphasizes the analysis of both feature-space and token-space covariance structures. Let the representation matrix at layer $l$ be defined as
\begin{equation}
\mathbf{X}_l \in \mathbb{R}^{(B \times T)\times D},
\end{equation}
where $B$ denotes the batch size, $T$ denotes the number of tokens, and $D$ denotes the embedding dimension. The feature covariance matrix is computed as

\begin{equation}
\mathbf{C}^{feat}_{l}
=
\frac{1}{N-1}
\left(
\mathbf{X}_{l}
-
\mathbf{1}\boldsymbol{\mu}_{l}^{T}
\right)^{T}
\left(
\mathbf{X}_{l}
-
\mathbf{1}\boldsymbol{\mu}_{l}^{T}
\right),
\end{equation}
where
\begin{equation}
N = B \times T
\end{equation}
and
\begin{equation}
\boldsymbol{\mu}_{l}
=
\frac{1}{N}
\sum_{i=1}^{N}
\mathbf{X}_{l}^{(i)}
\end{equation}
denotes the feature-wise mean representation.
The feature covariance matrix characterizes relationships between embedding dimensions and captures how variance is distributed across feature directions.

To investigate token interaction structure, a token covariance matrix is additionally computed as

\begin{equation}
\mathbf{C}^{token}_{l}
=
\frac{1}{D-1}
\left(
\mathbf{X}_{l}
-
\bar{\mathbf{X}}_{l}
\right)
\left(
\mathbf{X}_{l}
-
\bar{\mathbf{X}}_{l}
\right)^{T},
\end{equation}
where
\begin{equation}
\bar{\mathbf{X}}_{l}
=
\frac{1}{D}
\sum_{j=1}^{D}
\mathbf{X}_{l}^{(:,j)}
\end{equation}
denotes the token-wise mean representation. The token covariance matrix captures pairwise relationships between tokens and provides insight into token coupling and interaction structure throughout training.

It is important to note that the feature covariance matrix and token covariance matrix are related through the same underlying representation matrix. Let

\begin{equation}
\mathbf{A}
=
\mathbf{X}_{l}
-
\mathbf{1}\boldsymbol{\mu}_{l}^{T}.
\end{equation}
Then
\begin{equation}
\mathbf{C}^{feat}_{l}
\propto
\mathbf{A}^{T}\mathbf{A},
\end{equation}
while
\begin{equation}
\mathbf{C}^{token}_{l}
\propto
\mathbf{A}\mathbf{A}^{T}.
\end{equation}
From singular value decomposition,
\begin{equation}
\mathbf{A}
=
\mathbf{U}
\mathbf{\Sigma}
\mathbf{V}^{T},
\end{equation}
it follows that
\begin{equation}
\mathbf{A}^{T}\mathbf{A}
=
\mathbf{V}
\mathbf{\Sigma}^{2}
\mathbf{V}^{T}
\end{equation}
and
\begin{equation}
\mathbf{A}\mathbf{A}^{T}
=
\mathbf{U}
\mathbf{\Sigma}^{2}
\mathbf{U}^{T}.
\end{equation}

Consequently, both covariance matrices possess identical non-zero singular values and eigenvalue spectra, differing only in their associated eigenvectors. This relationship allows feature-space and token-space analyses to be interpreted as complementary views of the same underlying representation geometry.

\begin{figure*}[t]
    \centering

    \subfloat[TwoNN Intrinsic Dimension throughout training.\label{fig:twonn_mean}]{
        \includegraphics[width=0.48\textwidth]{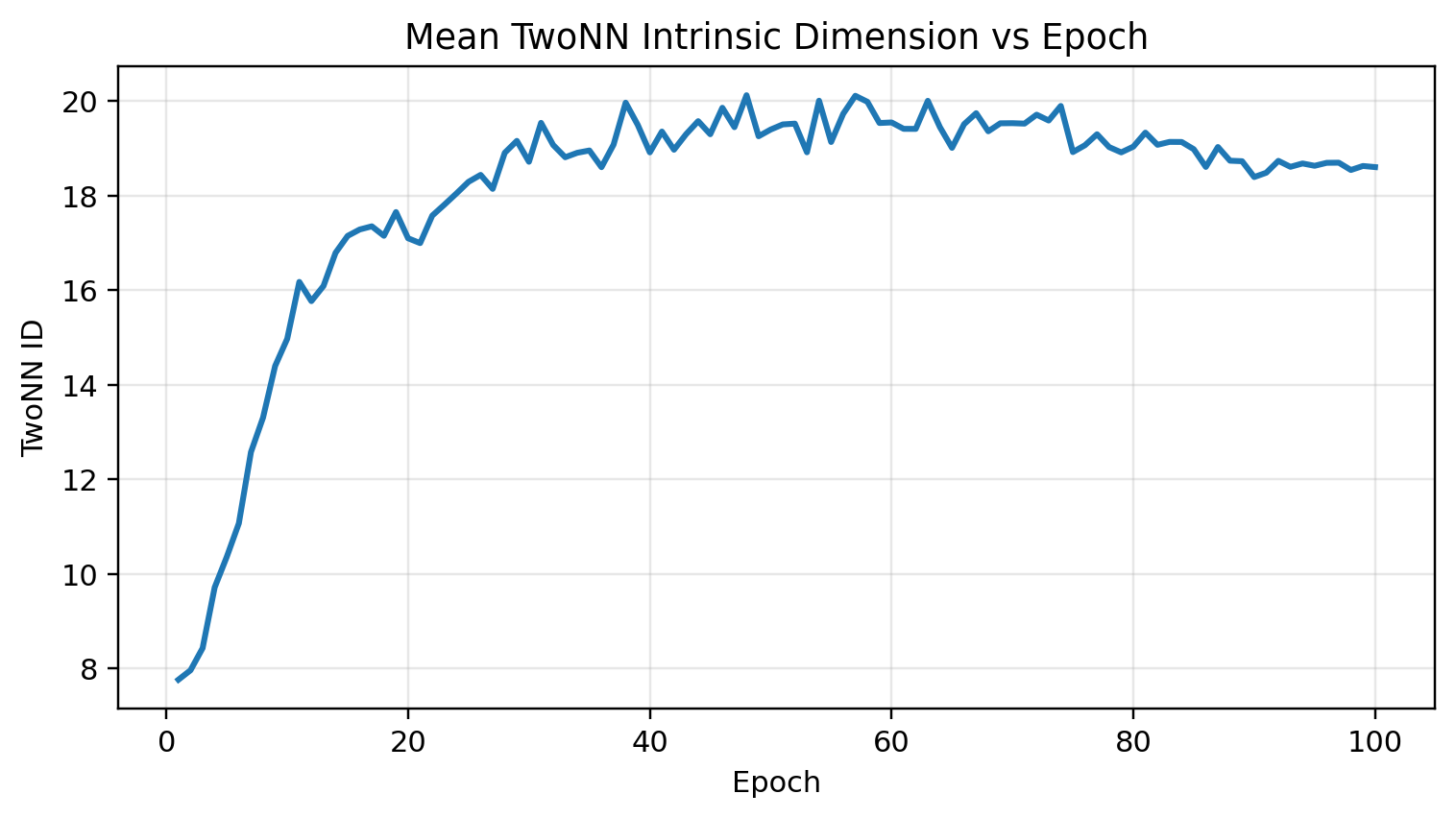}
    }
    \hfill
    \subfloat[Layer-wise TwoNN Intrinsic Dimension at Epoch 100.\label{fig:twonn_final}]{
        \includegraphics[width=0.48\textwidth]{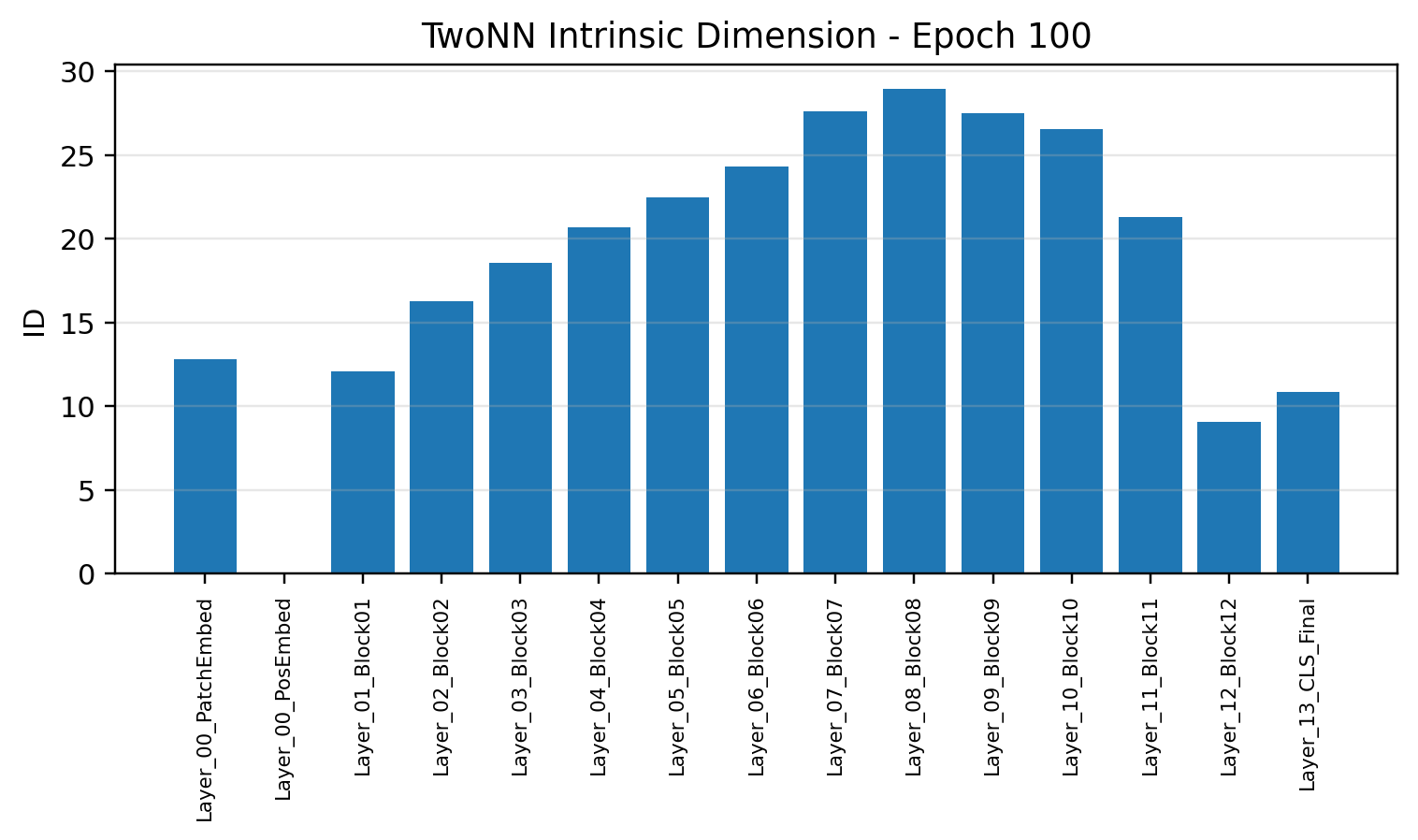}
    }

    \caption{Evolution of TwoNN intrinsic dimensionality. The left figure illustrates the evolution of intrinsic dimensionality across all Transformer layers throughout training, while the right figure summarizes the intrinsic dimensionality of each layer at convergence (Epoch 100). The middle Transformer blocks exhibit the highest intrinsic dimensionality, peaking at Layer 08 before gradually decreasing towards the final Transformer blocks.}
    \label{fig:twonn_results}
\end{figure*}

\section{Findings}
\label{sec:fin}

This section presents the principal observations obtained from TGO-II. The objective of this section is to report the observed characteristics of the learned representations without providing mechanistic explanations. Interpretation of these observations and the resulting hypotheses are deferred to the Discussion section. Unless otherwise stated, all measurements are computed on the fixed validation subset described in the Experimental Setup.

\subsection{SVCCA}

The evolution of the mean SVCCA throughout training is illustrated in Figure~\ref{fig:mean_svcca}, while Figure~\ref{fig:adj_svcca} presents the adjacent-layer SVCCA measured at the final epoch. A pronounced decrease in mean SVCCA is observed during the initial stages of training, followed by gradual stabilization during the later epochs. This trend is consistently observed across all Transformer layers. The adjacent-layer analysis further reveals that neighboring layers remain highly correlated at convergence. A noticeable reduction in adjacent-layer similarity occurs between Layers 4 and 5, after which the similarity progressively increases and remains consistently high for the remaining Transformer blocks.

\subsection{CKA}

Figure~\ref{fig:mean_cka} presents the evolution of the mean Centered Kernel Alignment (CKA) throughout training, while Figure~\ref{fig:adj_cka} illustrates the adjacent-layer CKA measured at the final epoch. CKA exhibits behaviour similar to SVCCA. A rapid decrease is observed during the early stages of optimization, followed by a gradual convergence to a stable value. This trend remains consistent across the complete training schedule. The adjacent-layer evaluation demonstrates that neighboring layers remain strongly aligned at convergence. The minimum similarity is observed between Layers 4 and 5, whereas Layers 6 through 12 exhibit consistently high representational similarity.

\subsection{Two-NN Intrinsic Dimensionality}
\label{subsec:id}
The evolution of intrinsic dimensionality throughout training is shown in Figure~\ref{fig:twonn_mean}, while the intrinsic dimensionality measured at the final epoch is presented in Figure~\ref{fig:twonn_final}. Across nearly all Transformer layers, the estimated intrinsic dimensionality increases rapidly during the initial stages of optimization before gradually converging to stable values. This behaviour is consistently observed throughout the training schedule. The final layer-wise intrinsic dimensionality reveals a distinct depth-dependent trend. Beginning from the early Transformer blocks, the intrinsic dimension increases progressively with network depth, reaching its maximum at Layer~08. Layers~07--10 consistently exhibit the largest intrinsic dimensions, indicating the highest local manifold complexity within the network. Beyond Layer~10, the intrinsic dimensionality decreases substantially through the final Transformer blocks, with Layer~12 exhibiting one of the lowest values among all Transformer layers. The final CLS representation exhibits a modest increase relative to Layer~12 while remaining below the peak values observed in the intermediate layers. As expected, the positional embedding exhibits negligible intrinsic dimensionality due to its deterministic nature and absence of sample-dependent variation.
\subsection{Token Covariance}
\label{subsec:tokcov}
\begin{figure*}[t]
    \centering
    \includegraphics[width=0.7\linewidth]{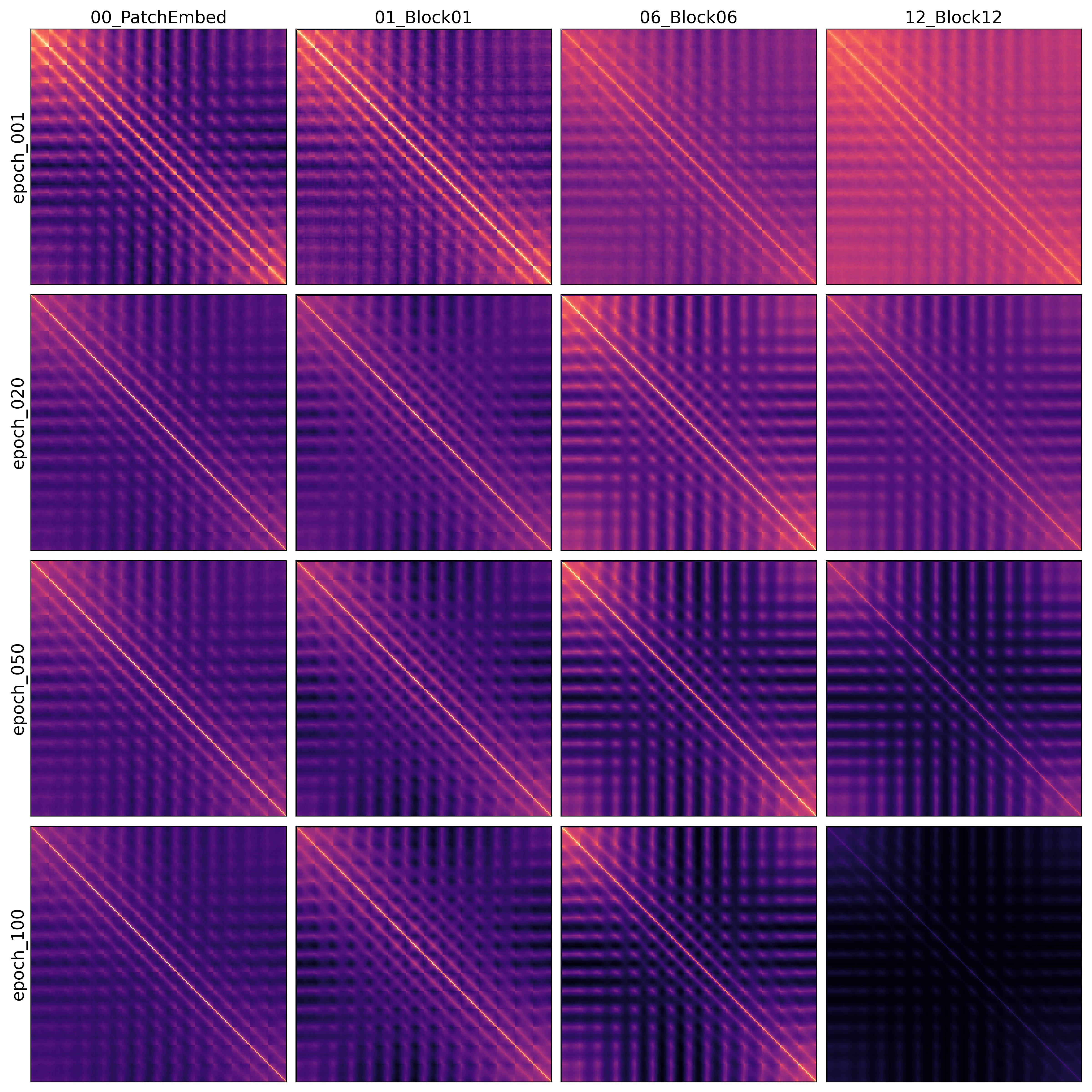}
    \caption{Token Covariance matrces for different layers and epochs}
    \label{fig:token_covariance}
\end{figure*}

Representative token covariance matrices are shown in Figure~\ref{fig:token_covariance}. Four representative layers (Patch Embedding, Block~01, Block~06 and Block~12) are visualized at Epochs 1, 20, 50 and 100 to illustrate the evolution of token interaction throughout training. The token covariance matrices exhibit highly structured interaction patterns throughout the optimization process. Rather than converging toward a diagonal covariance matrix, substantial off-diagonal covariance persists across all training epochs, indicating that statistical dependencies between tokens remain present throughout training. As optimization progresses, the covariance structures evolve from comparatively diffuse interaction patterns into increasingly organized stripe-like structures. This progression is consistently observed across all analyzed layers, although the magnitude of structural evolution varies with network depth. The Patch Embedding layer exhibits comparatively modest changes throughout training, whereas deeper Transformer blocks develop increasingly pronounced covariance structures. In particular, the final Transformer block (Layer~12) undergoes the most substantial structural evolution, producing the most organized interaction pattern by the end of training. 


\section{Hypotheses}
Section~\ref{sec:fin} discusses observations that TGO-II has thus produced in detail. When linked to the findings and hypotheses described in TGO-1~\cite{tgoi}, we can observe a few significant patterns. In TGO-I we had explored three major hypotheses revolving around \emph{Token Diversification}, \emph{Semantic Expansion}, \emph{Redundancy Removal}. One of the most important observation was \emph{Layer Clustering}, where after layer 4 there was a significant similarity in the effective rank and spectral dynamics evolution for layers 5 through 7. The following discussion lists hypotheses based on the observed trends.

\subsection{Hypothesis I: Token Diversification}
This is a hypothesis from TGO-I~\cite{tgoi} which claims that as training progresses the tokens spontaneously decouple leading to the significant directional exploration in spectral observables. However, as shown in Sub-section~\ref{subsec:tokcov} it is clear that the tokens do not decouple and attain complete independence and hence might not be considered as the sole reason for the directional exploration of the feature covariance matrix. What we can conclude is that \textbf{\emph{Token Diversification} does take place as the training progresses, however it might not be the only strong contributor to dimensional and directional expansion.}

\subsection{Hypothesis II: Semantic Expansion}

The Semantic Expansion hypothesis proposed in TGO-I~\cite{tgoi} suggests that the increase in spectral complexity observed during training arises from the progressive discovery and encoding of increasingly rich semantic concepts. TGO-II provides indirect evidence relevant to this hypothesis through representational similarity analysis. Both SVCCA and CKA demonstrate a consistent decrease throughout training, indicating that Transformer layers progressively specialize and learn increasingly distinct representations. However, representational specialization alone does not directly imply semantic expansion. The present observatory does not explicitly measure semantic information, class separation, or feature interpretability. Consequently, TGO-II neither confirms nor rejects the Semantic Expansion hypothesis. Instead, it identifies representational specialization as a potential prerequisite for semantic expansion, motivating future observatories involving semantic probing, class-separation analysis, and representation interpretability.

\subsection{Hypothesis III: Manifold Expansion Hypothesis}

The most significant observation obtained from TGO-II is the strong qualitative agreement between the evolution of TwoNN intrinsic dimensionality as shown in ~\ref{subsec:id} and the spectral expansion previously observed in TGO-I. Throughout training, intrinsic dimensionality increases rapidly before converging, closely mirroring the evolution of Effective Rank and other spectral observables measured on feature covariance matrices. This observation motivates the \emph{Manifold Expansion Hypothesis}: the increase in explored covariance directions is primarily enabled by the progressive expansion of the underlying representation manifold into higher intrinsic dimensions. As additional degrees of freedom become available, the network is capable of representing increasingly diverse feature relationships, naturally resulting in richer covariance spectra and higher effective ranks. Hence, TGO-II concludes that the \textbf{Semantic manifold expands to higher dimensions as the epochs progress}. Unlike the Token Diversification hypothesis, this explanation does not require tokens to become statistically independent. Instead, it attributes the observed spectral expansion to the increasing complexity of the learned representation manifold. While further mathematical analysis is required to establish a formal relationship between intrinsic dimensionality and covariance spectra, the empirical agreement between both observatories suggests that manifold expansion is a plausible explanation for the spectral evolution observed during Transformer training.

\subsection{Hypothesis IV: Transition Zone Hypothesis}

One of the most intriguing observations across both TGO-I and TGO-II is the consistent change in representational behaviour occurring around the fourth and fifth Transformer blocks. In TGO-I, Layers~5--8 exhibited remarkably similar spectral dynamics and Effective Rank evolution, suggesting the emergence of a distinct operational regime within the network. TGO-II further reveals that intrinsic dimensionality continues to increase beyond these layers while representational similarity progressively decreases, indicating increasing functional specialization. These observations motivate the \emph{Transition Zone Hypothesis}. Rather than viewing all Transformer blocks as performing homogeneous computations, we hypothesize that \textbf{the region around Layers~4--5 marks the transition between two qualitatively different stages of representation processing. Early layers may primarily transform low-level token interactions inherited from the patch embeddings, whereas subsequent layers increasingly operate on abstract and context-dependent representations.}

Importantly, this interpretation remains speculative. TGO-II does not directly measure semantic content or task-specific information encoded within the representations. Consequently, the proposed transition should be interpreted as a change in representational geometry rather than definitive evidence of semantic processing. Future observatories involving token trajectories, semantic probing, attention routing, and class-separation analysis are required to validate this hypothesis.

\section{Conclusion}

\emph{TGO-II: Representation Geometry Observatory} continues the Transformer Geometry Observatory (TGO) framework by analyzing the representational dynamics of Vision Transformers throughout supervised training. While TGO-I investigated the evolution of covariance spectra and feature-space geometry, TGO-II focused on representation similarity, intrinsic dimensionality, and token interaction structure. Using Centered Kernel Alignment (CKA), Singular Vector Canonical Correlation Analysis (SVCCA), Two-Nearest Neighbor Intrinsic Dimensionality (TwoNN-ID), and token covariance analysis, several consistent patterns were observed. Both CKA and SVCCA exhibited a progressive decrease throughout training, indicating that Transformer layers become increasingly specialized as optimization proceeds. Simultaneously, intrinsic dimensionality increased across the network before converging, suggesting that the learned representation manifold progressively occupies a richer set of locally accessible degrees of freedom. 

Furthermore, token covariance analysis demonstrated that strong token interaction structure persists throughout the entire training process, indicating that increasing representational complexity is not accompanied by complete token independence. These observations provide new insight into the geometric evolution of Transformer representations. In particular, the combination of increasing intrinsic dimensionality, decreasing representational similarity, and persistent token coupling suggests that representational complexity emerges through increasingly sophisticated transformations rather than through progressive token decoupling. The strong correspondence between the evolution of intrinsic dimensionality and the covariance-based spectral observables reported in TGO-I further motivates the hypothesis that manifold expansion provides a plausible geometric explanation for the observed increase in covariance rank and variance utilization during training. TGO-II also revisits several hypotheses proposed in TGO-I. The results indicate that complete token decoupling is unlikely to be the dominant mechanism responsible for covariance expansion, while the Semantic Expansion hypothesis remains inconclusive without direct measurements of semantic content. Additionally, the observed representational behaviour across network depth motivates the hypothesis that Transformer computation may undergo distinct geometric phases during training, with intermediate layers exhibiting characteristic representational organization that warrants further investigation.

Collectively, the findings of TGO-I and TGO-II suggest that Transformer learning is governed by coordinated geometric evolution across multiple complementary spaces. Rather than relying on a single observable, the Transformer Geometry Observatory demonstrates that covariance geometry, representation geometry, and manifold geometry together provide a richer understanding of how Vision Transformers organize and refine their internal representations throughout optimization.

\section{Future Work}

TGO-II extends the Transformer Geometry Observatory by investigating representation similarity, intrinsic dimensionality, and token interaction dynamics throughout Vision Transformer training. While several hypotheses proposed in TGO-I have been refined, the observations presented in this work motivate a new series of observatories aimed at understanding the mechanisms responsible for representation evolution.

\subsection{TGO-III: Semantic Geometry Observatory}

Although TGO-II demonstrates increasing representational specialization and manifold expansion, it does not directly measure semantic information. Future work will therefore investigate the emergence of semantic structure through linear probing, class-separation analysis, nearest-neighbor purity, Fisher discriminant analysis, and label alignment. These experiments aim to determine whether the increasing intrinsic dimensionality observed during training corresponds to the discovery of increasingly rich semantic representations.

\subsection{TGO-IV: Token Dynamics Observatory}

The persistence of strong token coupling throughout TGO-II suggests that representational complexity does not arise solely through token independence. Future observatories will therefore investigate token-level dynamics using cosine similarity, token trajectory visualization, token drift, local manifold analysis, neighborhood preservation, and temporal token evolution. These measurements aim to characterize how token interactions evolve despite maintaining strong covariance structure.

\subsection{TGO-V: Attention Geometry Observatory}

The transition-like behaviour observed between the early and intermediate Transformer blocks motivates a detailed investigation of attention dynamics. Future work will analyze attention entropy, head specialization, routing behaviour, token dominance, sparsity, and information flow throughout training. These experiments seek to determine whether changes in attention geometry coincide with the representational transition hypothesized in TGO-II.

\subsection{TGO-VI: Optimization Geometry Observatory}

Future work will investigate the optimization landscape governing Transformer learning through gradient geometry, Hessian spectra, curvature evolution, loss landscape topology, and optimization phase transitions. Understanding these dynamics may reveal whether the geometric behaviours observed by TGO-I and TGO-II emerge from identifiable optimization regimes.

\subsection{Long-Term Objective}

The long-term objective of the Transformer Geometry Observatory is to establish a unified geometric framework for understanding Transformer learning dynamics. By integrating covariance geometry, representation geometry, semantic geometry, token dynamics, attention geometry, and optimization geometry, future observatories aim to transform empirical observations into mechanistic explanations of Transformer computation.

Ultimately, this framework seeks to identify the geometric principles governing representation learning, reveal architectural bottlenecks and redundant computation, and provide theoretically grounded insights that can inform the design of future efficient Transformer architectures, adaptive sparse models, and neuromorphic learning systems.

\bibliographystyle{IEEEtran}
\bibliography{references}
\end{document}